\documentclass[11pt]{article}

\usepackage[final]{acl}

\usepackage{booktabs}
\usepackage{xcolor}
\usepackage{colortbl}
\usepackage{tabularx}
\usepackage{pifont}
\definecolor{triogray}{RGB}{245,245,245}
\definecolor{triored}{RGB}{205, 92, 92}
\definecolor{triogreen}{RGB}{34, 139, 34}
\definecolor{trioagent}{RGB}{0, 0, 128}

\usepackage{times}
\usepackage{latexsym}
\usepackage[T1]{fontenc}
\usepackage[utf8]{inputenc}
\usepackage{microtype} 
\usepackage{inconsolata} 
\usepackage{graphicx} 
\usepackage{setspace} 
\usepackage{pifont} 

\usepackage{amsmath}
\usepackage{amssymb}
\usepackage{mathtools}
\usepackage{amsthm}
\usepackage[mathcal]{euscript}

\usepackage{amsmath,amsfonts,bm}
\usepackage{relsize}

\def\eqref#1{equation~\ref{#1}}

\def\1{\bm{1}}

\DeclareMathAlphabet{\mathsfit}{\encodingdefault}{\sfdefault}{m}{sl}
\SetMathAlphabet{\mathsfit}{bold}{\encodingdefault}{\sfdefault}{bx}{n}

\usepackage[ruled,vlined]{algorithm2e}
\usepackage{enumitem}
\SetKwInput{KwRequire}{Require}
\SetKwInput{KwEnsure}{Ensure}
\DontPrintSemicolon

\usepackage{wrapfig} 
\usepackage{subcaption}   

\usepackage{array}
\usepackage{booktabs}
\usepackage{siunitx}
\usepackage{makecell}
\usepackage{tabularx}
\usepackage{ragged2e}
\newcolumntype{C}[1]{>{\centering\arraybackslash}m{#1}} 
\usepackage{nicematrix}

\usepackage[table,dvipsnames]{xcolor}
\definecolor{mydarkblue}{rgb}{0,0.4,0.8} 
\hypersetup{
    colorlinks=true,
    linkcolor=black,
    citecolor=mydarkblue,
    filecolor=magenta,      
    urlcolor=magenta,
}

\usepackage{placeins} 

\usepackage{cuted} 

\usepackage{dblfloatfix}

\newcommand{\VDCell}[2]{%
  \raisebox{-.5\height}{%
    \shortstack[c]{%
      #1\\[-0.3ex]{\scriptsize({\color{ForestGreen}{$\downarrow$}\,#2})}%
    }%
  }%
}
\newcommand{\VDCellR}[2]{%
  \raisebox{-.5\height}{%
    \shortstack[c]{%
      #1\\[-0.3ex]{\scriptsize({\color{BrickRed!85}{$\uparrow$}\,#2})}%
    }%
  }%
}

\newcommand{\MethodCell}[1]{%
  \raisebox{-.5\height}{%
    \shortstack[l]{
      #1\\[-0.3ex]{\scriptsize\vphantom{(0)}}
    }%
  }%
}

\title{
TRIO: Task Refinement with Individual Optimization for Retrieval-Augmented Generation
}
\def\method{\texttt{TRIO}} 

\author{
 \textbf{Yingtao Ren\textsuperscript{1}},
 \textbf{Xiao Luo\textsuperscript{2}},
 \textbf{Yiwei Fu\textsuperscript{3}},
 \textbf{Jinzhao Zhou\textsuperscript{1}},
 \textbf{Yu-Cheng Chang\textsuperscript{1}},
 \textbf{Chin-Teng Lin\textsuperscript{1}},
\\
 \textsuperscript{1}University of Technology Sydney,
 \textsuperscript{2}University of Wisconsin--Madison,
 \textsuperscript{3}Peking University
\\
 \small{
   \textbf{Correspondence:} \href{chin-teng.lin@uts.edu.au}{chin-teng.lin@uts.edu.au}
 }
}

\begin{document}
\maketitle

\begin{abstract}

This paper studies retrieval-augmented generation (RAG), which leverages external knowledge to improve the performance of language models. Despite the remarkable progress, current RAG approaches are still far from satisfactory due to inadequate retrieval and potential hallucinations. Towards this end, we propose a novel approach named \underline{T}ask \underline{R}efinement with \underline{I}ndividual \underline{O}ptimization (TRIO) for RAG. 
The core idea of our TRIO is to equip three well-designed agents with an iterative refinement paradigm by aligning their individual optimization objectives.
Specifically, we first expand ambiguous queries with a query extension agent and quantify the reward based on the improved answer quality, which can be utilized to refine our agent for higher expansion efficiency. To enhance the retrieval quality, we include a unanimity voting to annotate the expanded query as insufficient or sufficient, and their generative outcomes are utilized as ground truth to optimize the judge agent. To further improve evidence utilization and answer accuracy, an answer agent is optimized with dynamic matching-based rewards with curriculum learning for final outputs. Extensive experiments across multiple benchmark datasets validate the effectiveness of the proposed TRIO in comparison with state-of-the-art approaches. Our code is available at \url{https://github.com/yingtaoren/TRIO}.

\end{abstract}

\section{Introduction}

Large language models (LLMs) have achieved remarkable success in natural language understanding and reasoning. By pre-training on a series of high-quality instruction datasets, LLMs can acquire a wide range of factual knowledge~\citep{GPTreport,newsurvey}. These pre-trained models have enabled a wide range of applications beyond conventional language tasks~\citep{LLM4scienceL,fu2026zeus}. Nevertheless, pre-trained models may fail to capture the most up-to-date information in highly specialized domains, particularly those involving private or sensitive knowledge, such as healthcare and law~\citep{hallunicationllm}. This issue limits LLMs’ capabilities on some knowledge-intensive tasks, even leading to hallucinations.

\begin{figure}[t] 
  \centering
  \includegraphics[width=\linewidth]{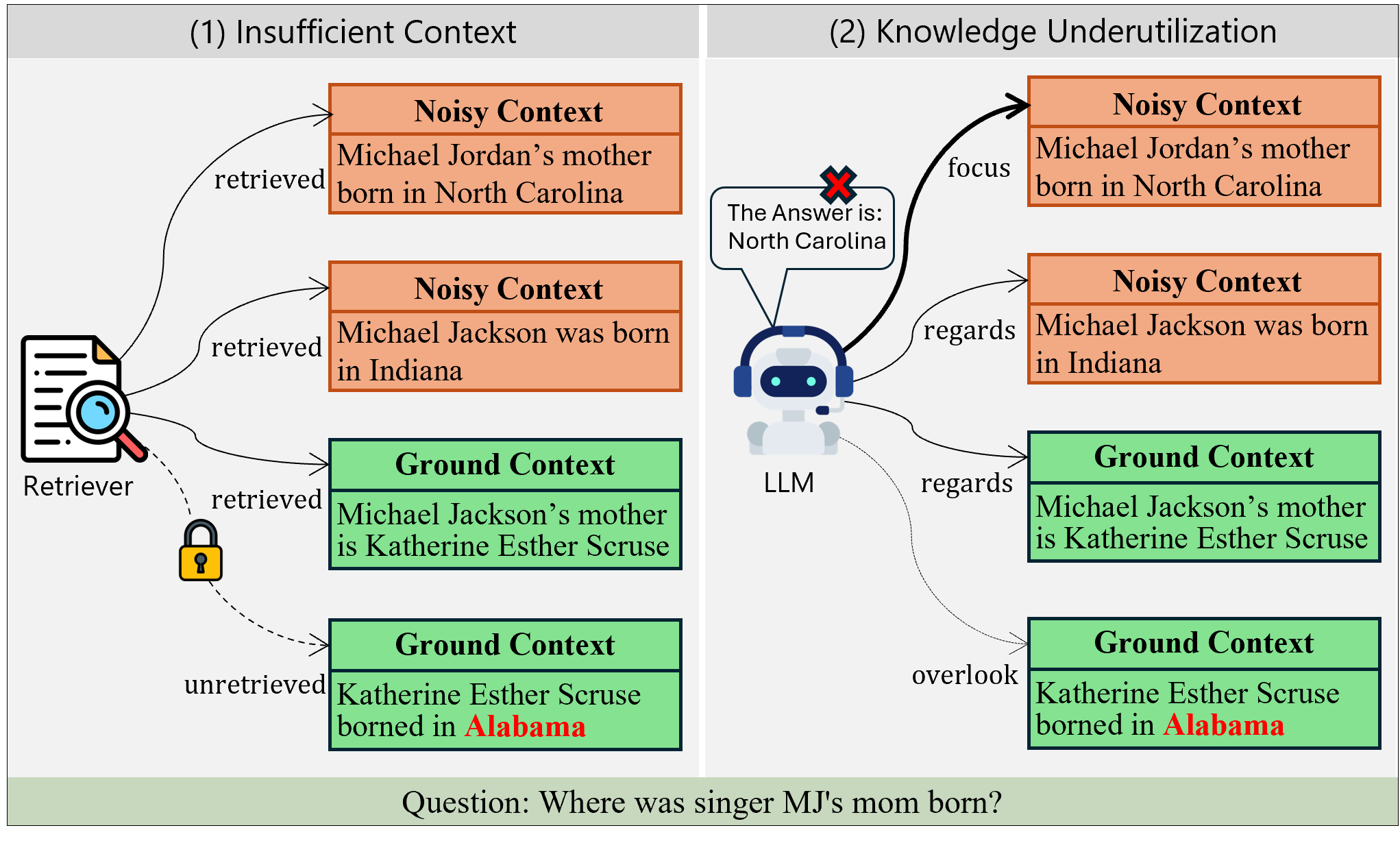}
  \captionsetup{skip=0pt}
  \caption{Problem illustration. RAG struggles with inefficient knowledge acquisition in the retrieval stage and knowledge underutilization in the answer stage.}
  \label{fig:problem_define}
\end{figure}

To enhance the response quality, RAG has been employed as a framework that leverages external knowledge to reduce reliance on the pre-trained LLMs’ parametric knowledge~\citep{wang2026comorag,gu2026evirag}. Typical RAG systems include two main steps: retrieval and generation. In the retrieval step, the key point is to locate the lacking knowledge and define a retrieval query to retrieve the knowledge from an external corpus. And then, in the generation stage, the retrieved contents are integrated with the original question and fed into the LLM. The augmented input provides the LLM with relevant knowledge and the latest information, thereby enhancing the quality of responses and reducing hallucinations~\citep{mishra2024finegrained}. The main advantage of RAG is that it provides external contextual information to answer questions when combined with the model’s parametric knowledge~\citep{scragL}. However, this advantage is constrained by two challenges. First, the system may not be able to accurately identify the missing knowledge and generate an effective query for retrieval. An inefficient query formulation may lead to the retrieval of numerous irrelevant or even biased documents, thereby increasing the model’s reasoning burden~\citep{autorag,cuconasu2024power}. Second, even when relevant documents are retrieved, the model may be unable to appropriately utilize the provided context, which may lead to further hallucinations~\citep{foundragacl,longcontext,AstuteRAG,dscan}.

Existing approaches to optimizing the pipeline of RAG include the branching method \citep{SuRe2404,2024replug}, context ranking \citep{yu2024rankrag}, and iterative retrieval \citep{asai2024selfrag,heetal2024retrieving}. These approaches merely adjust the pipeline of RAG, remaining insufficient to counteract the noise or bias introduced by retrieved knowledge. Furthermore, many recent works fine-tune specific LLMs for the RAG pipeline to improve robustness and factual consistency in the generation stage. Such efforts include supervised fine-tuning with human-annotated datasets \citep{2024ragtruth} and utilizing reinforcement learning to integrate reasoning ability with retrieval \citep{R1Searcher}. However, by placing the burden of generating retrieval queries and answering on the same backbone model, these approaches inherit the models’ intrinsic biases, limiting their robustness and ability to generate effective queries. 
 
 Recently, a growing trend in LLMs is to break complex processes into multiple specialized agents or modules that work \citep{chen2024agentverse,llmagentL} collaboratively. Instead of uniform models handling everything, each agent is responsible for a specific subtask in multi-agent systems~\citep{COPPER}. This paradigm has also been introduced into the RAG framework, resulting in a remarkable improvement in removing hallucinations and effectively utilizing external knowledge \citep{marag,drag}. However, existing multi-agent RAG frameworks still share the same backbone without adaptive adjustment, which potentially undermines response reliability.

To address these challenges, we propose a multi-agent system with Task Refinement and Individual Optimization (TRIO), an innovative multi-agent collaborative RAG system that tightly couples the retrieval and generation stages to improve retrieval and answer quality. Unlike existing multi-agent RAG approaches that utilize uniform backbones~\citep{drag}, TRIO assigns task-adaptive agents to distinct subtasks, allowing each agent to specialize in its role. This targeted specialization enables the system to pinpoint missing knowledge and utilize retrieved knowledge more effectively, yielding higher answer quality. Because each agent operates on a narrow slice of the pipeline, TRIO also lowers training cost and improves reliability.

Our TRIO implements an adaptive multi-agent RAG framework that decomposes the workflow into three subtasks: retrieval query generation, sufficient information judgment, and answer generation, each subtask handled by an optimized expert agent. This adaptive design enables effective handling of diverse questions and yields robust performance across domains. 
In addition, we present a streamlined data-generation and training protocol that tightly couples the optimization of all three agents, significantly improving effectiveness while maintaining modest supervision overhead.
Crucially, the training data construction and fine-tuning procedures embed substantial cross-agent collaboration, which strengthens coordination within the multi-agent system and tightly couples the subtasks of the pipeline. The protocol facilitates a straightforward transfer to a wide range of RAG scenarios and provides a practical path to enhance real-world RAG deployments. Our main contributions can be summarized as follows:
\vspace{-0.5em}
\begin{enumerate}[label=(\arabic*),leftmargin=20pt,labelindent=0pt,itemindent=0pt,align=left,labelsep=0.1em]
\setlength\itemsep{0em}
 
\item Existing RAG systems struggle with inefficient knowledge acquisition and hallucination issues. We address these challenges through TRIO, which strategically assigns specialized agents to distinct RAG pipeline stages.
\item Our TRIO introduces novel data generation methods that expand limited supervised data to align across different subtasks of RAG and further adopts adaptive fine-tuning paradigms for each agent. This design helps enhance retrieval efficiency and answer quality.
\item Comprehensive experiments across the latest benchmarks demonstrate that our TRIO outperforms a range of competitive baselines. In addition, the framework can be readily extended to various RAG scenarios. 
\end{enumerate}

\begin{figure*}[!t]
    \centering
    \setlength{\abovecaptionskip}{0.1pt}
    \includegraphics[width=\textwidth]{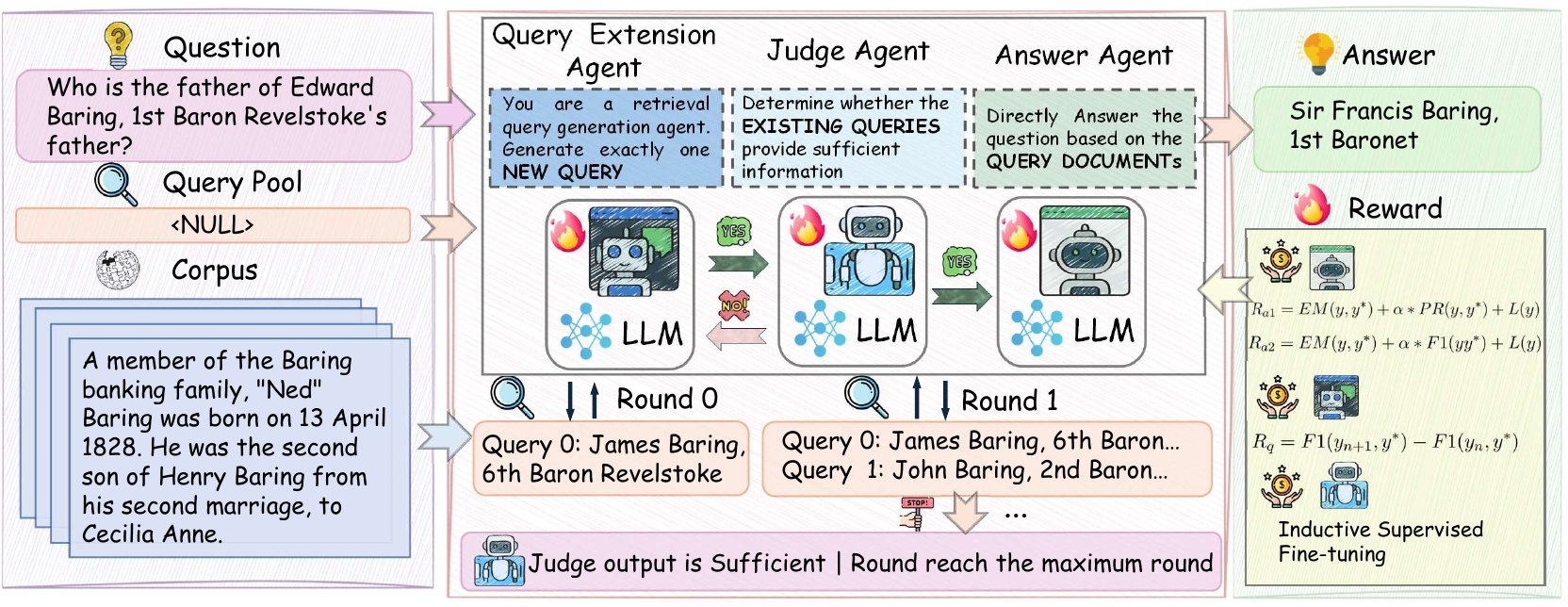} 
    \caption{An overview of our TRIO framework. The tight cooperation among three specific optimized agents improves the retrieval efficiency and answer quality.}
    \label{fig:framework}
    \vspace{-15pt}
\end{figure*} 

\section{Related Work}

\textbf{Retrieval-Augmented Generation.} Large language models have achieved strong performance across diverse tasks~\citep{naturehealth}. However, they remain susceptible to hallucinations and insufficient context. Retrieval augmented generation addresses this limitation by retrieving external evidence and conditioning the model’s output on the retrieved context~\citep{chen2025scirerankbench,fan2024ragsurvey}. Early RAG pipelines use the original input as a query to search on an external corpus, and then integrate the retrieval context into the model’s input as supplementary knowledge~\citep{RAGsurvey}. To meet more complex retrieval needs, many methods introduce iterative retrieval generation branching~\citep{SuRe2404} or loops~\citep{retrobust, rqrag, flare,ITRG}, letting the model adapt its retrieval strategy when evidence is insufficient. With the rise of chain-of-thought prompting~\citep{enhancecot}, reasoning traces have also been used to improve retrieval efficiency and recall~\citep{reasoningRAGL,IRCOT}. Some multi-agent methods have also been explored in RAG to enhance retrieval quality~\citep{drag,marag}. 
Despite these advances, most existing methods still share a single backbone model across all roles in the RAG pipeline and lack targeted fine-tuning, which limits further performance gains.\\
\textbf{Multi-Agent System.} Recent work shows that several agents' interactions can substantially enhance LLM capabilities across three main regimes~\citep{agentsurveyacl}. First, debate-based frameworks run adversarial dialogues or critiques in which agents challenge each other to refine reasoning and factuality~\citep{madicml,zhang2025multillmagents}. Second, ensemble coordination runs multiple agents in parallel with minimal communication and aggregates their outputs via voting, validation, or game-theoretic consensus~\citep{wang2025mixtureofagents,ensembleagentaaai}. Third, role-specialized collaboration assigns complementary roles to decompose problems and iteratively refine solutions~\citep{plangenllmsacl,liu2025pcagent}. However, the multi-agent setting for RAG remains underexplored. Contemporary RAG pipelines typically rely on a single LLM to both retrieve and generate, with limited support for collaboration among specialized agents.

\vspace{-4pt}
\section{The Proposed TRIO} \label{sec:method}
\vspace{-2pt}

\textbf{Problem Definition.} Retrieval augmented generation enhances large language models by integrating external knowledge as context that is not stored in the model’s parameters. Given a collection of retrieval text chunks $C$ and a corresponding retrieval query $Q$, which was identified by the LLM to obtain knowledge from an external corpus. If necessary, it can perform multiple retrieval rounds to acquire additional knowledge. However, excessive context may introduce noise and increase the risk of hallucination.
Our objective is to build an efficient multi-agent RAG system, where each agent is specialized and fine-tuned to a specific subtask within the RAG pipeline. 
This multi-agent RAG system must operate as a tightly coupled loop, collaboratively formulating retrieval queries targeting missing evidence in an external corpus to answer question $x$ while minimizing irrelevant content, and then seamlessly utilizing the acquired context to generate accurate responses.
The central problem is training multiple agents to cooperate across retrieval, judgment, and answer generation so they function as an effective whole.

\subsection{Framework Overview} \label{sec:overview}
We address a limitation of existing RAG systems that rely on uniform LLMs across all stages, leading to weak coordination among components. In TRIO, specialized agents are trained to adapt to their respective subtasks while engaging in substantial cross-agent cooperation. This training strategy strengthens coordination within the multi-agent RAG system, tightly couples the subtasks, and improves robustness and factual consistency. In particular, during the construction of training data and the training process, the query agent is trained to retrieve evidence aligned with the preferences of the answer agent, making it easier to produce accurate and unbiased answers. The judge agent is trained to understand the other agents’ behaviors and capabilities, enabling more reliable decisions about whether the current retrieval evidence is sufficient. The overall optimization loop is summarized in Algorithm~\ref{alg:maps}.

\subsection{Answer Agent Optimization with Matching-based Reinforcement Learning} \label{sec:answer_agent}

The answer agent is responsible for using the retrieved content to answer questions with high accuracy. The primary goal for training the answer agent is to enhance the utilization of retrieval content and avoid hallucinations. To train the agent, we build a training corpus that spans different numbers of retrieval steps and various chunk set sizes, enabling the agent to learn to ignore biases in the context and focus on the key evidence.

We sample training questions for the answer agent from multi-hop QA datasets, assigning difficulty labels (easy/medium/hard) and retaining medium and hard questions following~\citep{R1Searcher}. We then construct a retrieval-only debate pipeline~\citep{drag} that iteratively refines retrieval queries and retrieved contents, capping the debate at three rounds.
We store the query pool and retrieved chunks at the end of each debate round, treating each snapshot as a training sample. For each question, we generate multiple training samples per question by varying the number of collected queries and chunks. This setup trains the agent to filter noise and focus on key evidence across different context sizes. Further details on the debate pipeline are provided in Appendix~\ref{sec:data_generation}.

To improve the generative capabilities of the answer agent, we propose a two-stage outcome-based reinforcement learning (RL) training method based on the GRPO algorithm~\citep{deepseekGRPO}. The model is first trained to utilize the provided information and minimize internal hallucinations, using questions with the top 3 retrieval chunks for each retrieval query. Specifically, an outcome-based reward consists of an exact match (EM) score and a precision score, which are used in this stage. The EM is defined as follows:
\vspace{-0.5em}
\begin{equation}
EM(y,y^*) =
\begin{cases}
1, & \text{if the $y = y^*$} \\
0,   & \text{else},
\end{cases}
\label{eq:EM}
\end{equation}
where $y$ is the predicted answer generated by the answer agent and $ y*$ is the gold answer of the input question $x$. The precision score is:
\begin{equation}
PR(y,y^*) = \frac{IN(y,y^*)}{PN(y)},
\end{equation}
where the $PN(y)$ represents the word count of the answer, $IN$ indicates the word count of the intersection between the two answers. To further discourage explanations and other extraneous content in the answer, we add a format term to the reward:
\begin{equation}
L(y) = 
\begin{cases}
-1, & \text{if the length of $y$ larger than $l$} \\
0,   & \text{else},
\end{cases}
\end{equation}
where $l$ is a tuned hyper-parameter. Therefore, the final reward $R_{a1}$ of the stage is the sum of the outcome metrics and format metrics:
\begin{equation}
R_{a1} = EM(y,y^*) + \alpha * PR(y,y^*) + L(y),
\label{eq:reward1}
\end{equation}
where $\alpha$ is the tuned hyper-parameter.

Subsequently, the answer agent is trained to effectively locate key evidence in richer contexts and to explore broader reasoning paths, using questions with the top 5 retrieval chunks for each retrieval query. The outcome-based reward comprises an EM score and an F1 score:
\begin{equation}
R_{a2} = EM(y,y^*) + \alpha * F1(y,y^*) + L(y),
\label{eq:reward2}
\end{equation}
where the token-level $F1$ score is:
\begin{equation}
F1(y,y^*) = \frac{2 * IN(y,y^*)}{PN(y) + PN(y^*)}.
\label{eq:F1}
\end{equation}

\newcolumntype{Y}{>{\RaggedRight\arraybackslash}X}
\arrayrulecolor{black!55}          
\setlength{\arrayrulewidth}{0.4pt} 
\setlength{\tabcolsep}{6pt}
\renewcommand{\arraystretch}{1.18}

\newcommand{\Thought}[2]{\textbf{Thought #1:} #2\par}
\newcommand{\Action}[2]{\textbf{Action #1:} {\ttfamily\color{green!45!black}#2}\par}
\newcommand{\Obs}[2]{\textbf{Observation #1:} #2\par}
\newcommand{\Note}[1]{\textit{\color{black!60}#1}}

\subsection{Query Extension Agent Optimization with Outcome-based Reward} \label{sec:query_agent}

The goal of retrieval-query generation in RAG is to identify gaps in the current context and generate new retrieval queries, reducing whole-context bias and thereby providing the key evidence needed by the answer agent. Existing methods struggle with a lack of a clear signal to judge the quality of a single query. In TRIO, we train an adaptive query extension agent to improve retrieval quality based on the answer agent's generation quality.

We first use each original question as a retrieval query to retrieve the corresponding training sample, resulting in a query-refinement dataset. Utilizing the strong answer agent in Section~\ref{sec:answer_agent}, we define the quality of an individual query by the change in answer quality after adding the chunk and before. As with the answer agent, we train the query-extension agent using GRPO, with the query quality reward:
\begin{equation}
R_q = F1(y_{n+1},y^*) - F1(y_{n},y^*),
\label{eq:reward_q}
\end{equation}
where $y_n$ is the answer generated by the answer agent with chunks retrieved in the first $n$ rounds.
The training encourages the query extension agent to pinpoint missing evidence under varied contexts and to generate effective queries accordingly.

\subsection{Judge Agent Optimization with Inductive Supervised Fine-tuning} \label{sec:judge_agent}

To decide whether the query extension agent has retrieved sufficient evidence and when to stop retrieval and begin answering, thereby limiting hallucinations from long contexts. The judge agent estimates whether the retrieved context is sufficient for the current answer agent to answer correctly. 

As with the query extension agent, we use the answer agent’s response quality as the sole supervision signal for judging context quality. Unlike the query extension agent, the judge is trained via supervised fine-tuning~\citep{sufficient2025}. For training data generation, we also sample questions from multi-hop QA datasets and use the same Debate-RAG method with Section~\ref{sec:answer_agent} to generate judge training samples. In order to get the supervised signal for the judge agent, we follow the test-time labeling method in \citep{ttrl}, run the answer agent on each sample, and draw 16 stochastic predicted answers. If all 16 are correct, we label the sample as ‘Sufficient Information’; if all 16 are incorrect, we label it ‘Insufficient Information’. This procedure yields reliable supervised signals, helping the judge learn to assess context sufficiency across various domains and context lengths. Finally, we combine these labeled samples to form the judge enhancement dataset, used to train the judge agent via supervised fine-tuning. 

\begin{algorithm}[t]

\SetKwInput{KwRequire}{\hspace*{-0.7em}Require}
\SetKwInput{KwEnsure}{\hspace*{-0.7em}Ensure}

\caption{Training Algorithm of \method}
\label{alg:maps}

\setstretch{1.1} 
\KwRequire{Question dataset $Q$, Retriever $\mathcal{R}$, Corpus $C$, Backbone model $L$}
\KwEnsure{Query-Extension $\mathcal{A}_{q}$, Judge $\mathcal{A}_{j}$, Answer $\mathcal{A}_{g}$}

\BlankLine 

\nl Extract $Q_{1}$ (hard questions) and $Q_{2}$ (normal questions) from $Q$\;
\nl Conduct the retrieval debate method with $\mathcal{R}$ and $C$ on $Q_{1}$, build training dataset $T_1$\;
\nl Train $L$ on $T_1$ with two-stage RL to obtain $\mathcal{A}_{g}$, using reward functions in Eq.~\ref{eq:reward1} and Eq.~\ref{eq:reward2}\;
\nl Conduct a single retrieval using question on $Q_{2}$ with $\mathcal{R}$ and $C$, build training dataset $T_{2}$\;
\nl Train $L$ on $T_2$ with GRPO to obtain $\mathcal{A}_{q}$, using the reward in Eq.~\ref{eq:reward_q}\;
\nl Use $\mathcal{A}_{g}$ to answer all questions in $T_2$ with 16 samples, constructing training dataset $T_3$\;
\nl Supervised fine-tuning of $L$ on $T_3$, obtaining the judge agent $\mathcal{A}_{j}$.\;

\end{algorithm}

\begin{table*}[!ht]
\centering
\captionsetup{skip=5pt}
\setlength{\tabcolsep}{9pt} 
\setlength{\aboverulesep}{0pt}
\setlength{\belowrulesep}{0pt}
\renewcommand{\arraystretch}{1.05} 

\begin{tabular}{p{2.1cm} S[table-format=1.4] S[table-format=1.4] S[table-format=1.4] S[table-format=1.4] S[table-format=1.4] S[table-format=1.4]}
\toprule
~ & \multicolumn{2}{c}{\textbf{In-Domain}} & \multicolumn{3}{c}{\textbf{Out-of-Domain}} & \\
\cmidrule(lr){2-3} \cmidrule(lr){4-6}

\textbf{Method} & {\textbf{HotpotQA}} & {\textbf{2Wiki}} & {\textbf{NQ}} & {\textbf{Musique}} & {\textbf{Bamboogle}} & {\textbf{AVG}} \\
\midrule


\cmidrule(l{0pt}r{0pt}){1-7}

\rowcolor{gray!7}
\multicolumn{7}{c}{\textit{Single-Hop Retrieval}} \\
\cmidrule(l{0pt}r{0pt}){1-7}
Qwen3-4B        & 0.2770 & 0.1584 & 0.3357 & 0.0571 & 0.2240 & 0.2074 \\
Llama3.1-8B     & 0.3095 & 0.1639 & 0.3720 & 0.0533 & 0.1600 & 0.2237 \\
Qwen3-30B       & 0.2979 & 0.2479 & 0.3175 & 0.1278 & 0.2400 & 0.2606 \\
\cmidrule(l{0pt}r{0pt}){1-7}

\rowcolor{gray!7}
\multicolumn{7}{c}{\textit{Multi-Hop Retrieval}} \\
\cmidrule(l{0pt}r{0pt}){1-7}
SuRe            & 0.2257 & 0.1166 & 0.3532 & 0.0653 & 0.1680 & 0.1757 \\
IRCOT           & 0.1841 & 0.1266 & 0.2171 & 0.0690 & 0.2080 & 0.1505 \\
Iter-RetGen     & 0.3350 & 0.1651 & 0.3343 & 0.0724 & 0.1920 & 0.2282 \\
DRAG            & 0.3060 & 0.2094 & 0.3574 & 0.1171 & 0.2880 & 0.2490 \\
Self-RAG        & 0.1596 & 0.1159 & 0.3457 & 0.0422 & 0.0640 & 0.1530 \\
Search-R1       & 0.4116 & 0.4301 & \colorbox{red!15}{0.4382} & \colorbox{blue!15}{0.1836} & \colorbox{blue!15}{0.4640} & \colorbox{blue!15}{0.4034} \\
R1-Searcher     & 0.4012 & 0.4046 & 0.3681 & 0.1779 & \colorbox{blue!15}{0.4640} & 0.3779 \\


\cmidrule(l{0pt}r{0pt}){1-7}
TRIO-4B         & \colorbox{blue!15}{0.4174} & \colorbox{blue!15}{0.4373} & 0.3878 & 0.1547 & \colorbox{blue!15}{0.4640} & 0.3988 \\
TRIO-8B         & \colorbox{red!15}{0.4350} & \colorbox{red!15}{0.4710} & \colorbox{blue!15}{0.4339} & \colorbox{red!15}{0.1857} & \colorbox{red!15}{0.4960} & \colorbox{red!15}{0.4294} \\
\bottomrule
\end{tabular}
\caption{The overall evaluation results of TRIO and other baselines on five benchmarks. \colorbox{red!15}{Red} marks the best-performing method, and \colorbox{blue!15}{blue} represents the second-best-performing method. All methods are evaluated under the same settings. Our TRIO achieves the strongest performance.}
\label{tab:main_result}
\vspace{-10pt}
\end{table*}

\subsection{Inference Pipeline}
With well-designed training, each agent within TRIO coordinates closely to generate concise outputs. At inference time, the three agents collaborate to acquire external knowledge through two main steps: the retrieval step and the answer step. Standard RAG methods often suffer from insufficient or irrelevant retrieval chunks in the retrieval process. To address this, TRIO tightly couples the agents to ensure precise cooperation in an iterative retrieval process. Starting from an initial retrieval, the query agent uses the current context to identify knowledge gaps and generate targeted new queries, while the judge agent dynamically judges whether the accumulated information is sufficient to halt the iteration. This precise cooperation ensures the finalized query pool and retrieved chunks are highly concise. Once sufficient information is gathered, we switch to the answer step in which the answer agent generates the final answer based exclusively on the current context.

As the query and judge agents collaborate effectively to supply only the strictly necessary evidence, the answer agent is relieved from processing massively irrelevant contexts. Consequently, TRIO can generate highly concise output tokens to reach an accurate conclusion, mitigating hallucination risks and significantly improving overall inference efficiency. The detailed inference loop and agent prompts are provided in Appendix~\ref{sec:prompt_TRIO}.

\section{Experiment}

\subsection{Experimental Settings} \label{sec:experiment_settings}
\textbf{Baselines}. 
We implement base RAG with different generation models~\citep{flashrag,qwen3}, which represents the basic RAG method without optimization. Moreover, we compare TRIO with several recent RAG methods, including branching method SuRE~\citep{SuRe2404}, loop method Self-RAG~\citep{asai2024selfrag}, IRCoT~\citep{IRCOT}, and Iter-RetGen~\citep{ITRG}, R1 style reasoning RAG method Search-R1~\citep{searchR1COLM}, R1-Searcher~\citep{R1Searcher}, and multi-agent debate method DRAG~\citep{drag}.\\
\textbf{Datasets \& Evaluation metrics}. 
Evaluation experiments are conducted on the dev or test datasets of five benchmarks:
Hotpotqa~\citep{hotpotqa}, 2Wiki~\citep{2wiki}, NQ~\citep{NQ_benchmark}, Musique~\citep{musique}, and Bamboogle~\citep{bamboogle}. Hotpotqa and 2Wiki are in-domain datasets because parts of their training sets are used to train our agents and other training-based baselines. NQ, Musique, and Bamboogle are out-of-domain datasets used solely for evaluation. For evaluation metrics, we report the exact match score in the results. \\
\textbf{Implementation Details}. 
All compared models are reproduced and evaluated using FlashRAG~\citep{flashrag}. We employed E5-base-v2~\citep{E5base} as the retriever, and the retrieval corpus using the Wikipedia dump 2018~\citep{wikidateset}. For each question, the external evidence is capped at 20 retrieved chunks.
In the training processes of our TRIO, the backbone models are Llama3.1-8B-Instruct for TRIO-8B and Qwen3-4B-Instruct for TRIO-4B. We select questions from the training sets of Hotpotqa and 2Wiki to generate training samples. The trained models for the three agents are available in our code. More details are provided in Appendix~\ref{sec:prompt_TRIO}.

\begin{table*}[!ht]
\captionsetup{skip=5pt}
\centering
\label{tab:reward_design}
{\renewcommand{\arraystretch}{1.3}
\setlength{\aboverulesep}{0pt}
\setlength{\belowrulesep}{0pt}
\setlength{\cmidrulekern}{0pt}
\setlength{\tabcolsep}{7.5pt} 

\begin{tabular}{l
  C{8.6mm} C{8.8mm}  
  C{8.8mm} C{8.8mm}  
  C{8.8mm} C{8.8mm}  
  C{8.8mm} C{9.2mm}  
}
\toprule
 & \multicolumn{2}{c}{\textbf{Hotpotqa}} & \multicolumn{2}{c}{\textbf{2Wiki}} & \multicolumn{2}{c}{\textbf{Musique}} & \multicolumn{2}{c}{\textbf{Bamboogle}} \\
\cmidrule(l{2pt}r{2pt}){2-3}\cmidrule(l{2pt}r{2pt}){4-5}\cmidrule(l{2pt}r{2pt}){6-7}\cmidrule(l{2pt}r{2pt}){8-9}
\textbf{Variants} & \textbf{EM} & \textbf{F1} & \textbf{EM} & \textbf{F1} & \textbf{EM} & \textbf{F1} & \textbf{EM} & \textbf{F1} \\
\midrule
\MethodCell{TRIO w/o Query Opt.} &
\VDCell{0.4235}{2.64\%} &
\VDCell{0.5555}{3.73\%} &
\VDCell{0.4547}{3.46\%} &
\VDCell{0.5191}{8.29\%} &
\VDCell{0.1357}{26.93\%} &
\VDCell{0.2456}{12.85\%} &
\VDCell{0.4560}{8.06\%} &
\VDCell{0.5590}{9.84\%} \\
\MethodCell{TRIO w/o Judge Opt.} &
\VDCell{0.4284}{1.52\%} &
\VDCell{0.5587}{3.17\%} &
\VDCellR{0.4765}{1.17\%} & 
\VDCell{0.5416}{4.31\%} &
\VDCell{0.1593}{14.22\%} &
\VDCell{0.2635}{6.49\%} &
\VDCell{0.4480}{9.68\%} &
\VDCell{0.5643}{8.98\%} \\
\MethodCell{TRIO w/o Answer Opt.} &
\VDCell{0.3735}{14.14\%} &
\VDCell{0.4926}{14.63\%} &
\VDCell{0.2778}{41.02\%} &
\VDCell{0.3618}{36.08\%} &
\VDCell{0.1051}{43.40\%} &
\VDCell{0.1855}{34.17\%} &
\VDCell{0.2800}{43.55\%} &
\VDCell{0.3967}{36.02\%} \\

\cmidrule(l{0pt}r{0pt}){1-9}
\rowcolor{gray!15}
TRIO (Full)                & 0.4350 & 0.5770 & 0.4710 & 0.5660 & 0.1857 & 0.2818 & 0.4960 & 0.6200 \\
\bottomrule
\end{tabular}}

\caption{Ablation study results on the four multi-hop benchmarks with three variants.}
\label{tab:ablation_study}
\vspace{-12pt}
\end{table*}

\subsection{Empirical Results and Analysis} \label{sec:Empirical_result}
We conduct comprehensive experiments to evaluate the performance of TRIO against various baselines across five benchmark datasets. As reported in Table~\ref{tab:main_result}, TRIO-8B achieves strong and consistent improvements on four benchmarks, demonstrating its robustness in settings that demand complex retrieval and multi-step inference. TRIO-4B remains competitive with R1-style reasoning-based RAG methods while using a substantially smaller backbone model. On average, TRIO-8B improves 6.45\% over the second-best method. This superior performance can be attributed to TRIO’s decomposition of the RAG pipeline into smaller subtasks and to performing each with a well-designed agent, which makes both retrieval and answering more effective. In particular, compared to the other multi-agent baseline DRAG, our method TRIO-8B improves 72.45\%, indicating that targeted fine-tuning for each agent design substantially boosts multi-agent RAG. We also observe that TRIO-8B achieves strong results on the Musique and Bamboogle datasets without any training on these benchmarks. This suggests that the agents learn effective retrieval and coordination under our diverse training scheme, yielding robust performance on new test sets. Furthermore, while TRIO-4B outperforms baseline models on in-domain benchmarks, it degrades on out-of-domain datasets. This observation seemingly highlights the impact of scaling laws in RAG tasks, indicating that limited model capacity inherently limits the system's generalization capabilities. We further evaluate TRIO's robustness to noisy irrelevant contexts under different noise levels, with
detailed results and analysis provided in Appendix~\ref{sec:noise_robustness}.

\begin{figure*}[t] 
  \centering
  \begin{subfigure}[t]{0.33\textwidth}
    \centering
    \includegraphics[width=\linewidth]{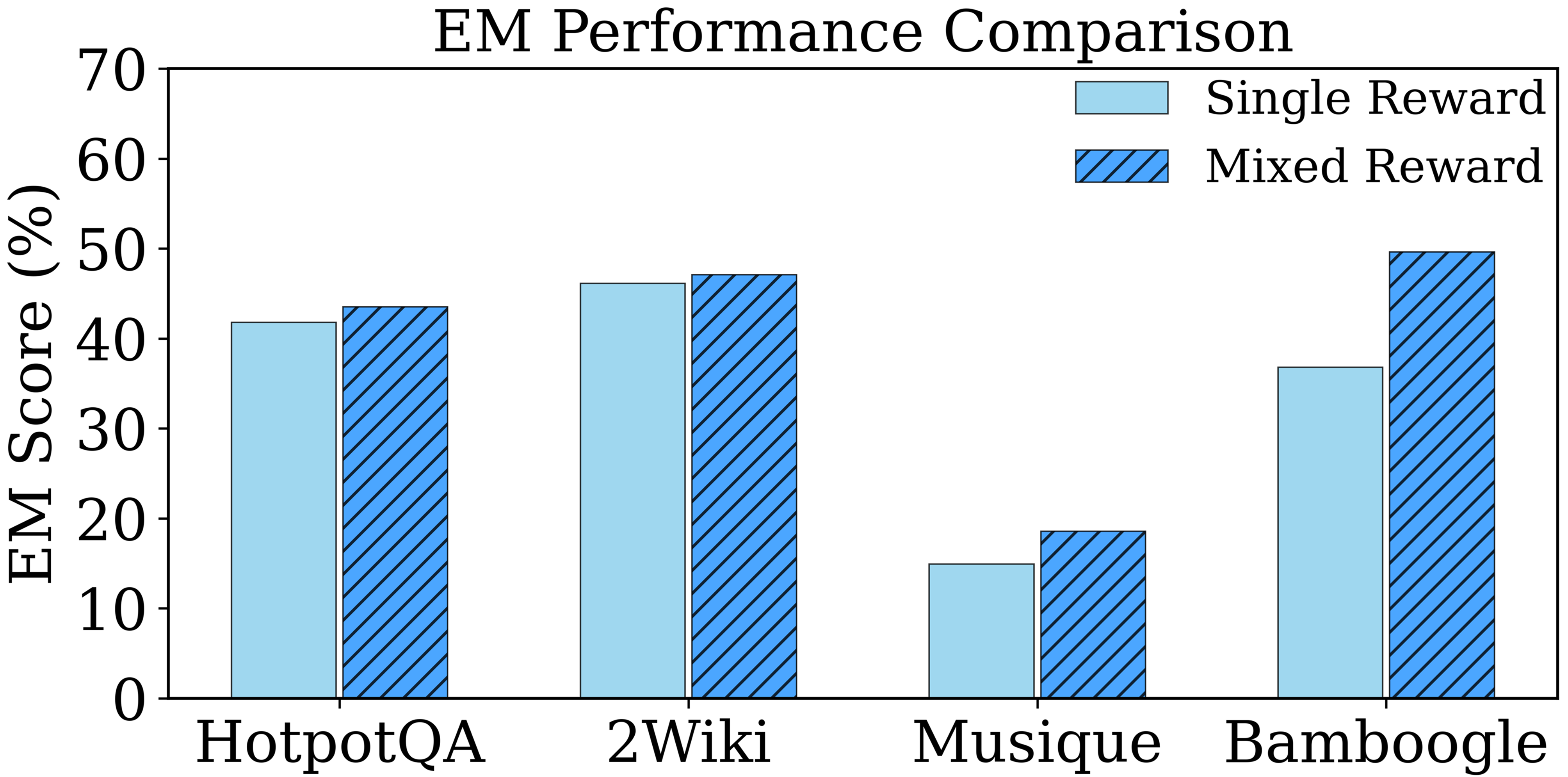}
    \caption{EM}
    \label{fig:three:a}
  \end{subfigure}%
  \begin{subfigure}[t]{0.33\textwidth}
    \centering
    \includegraphics[width=\linewidth]{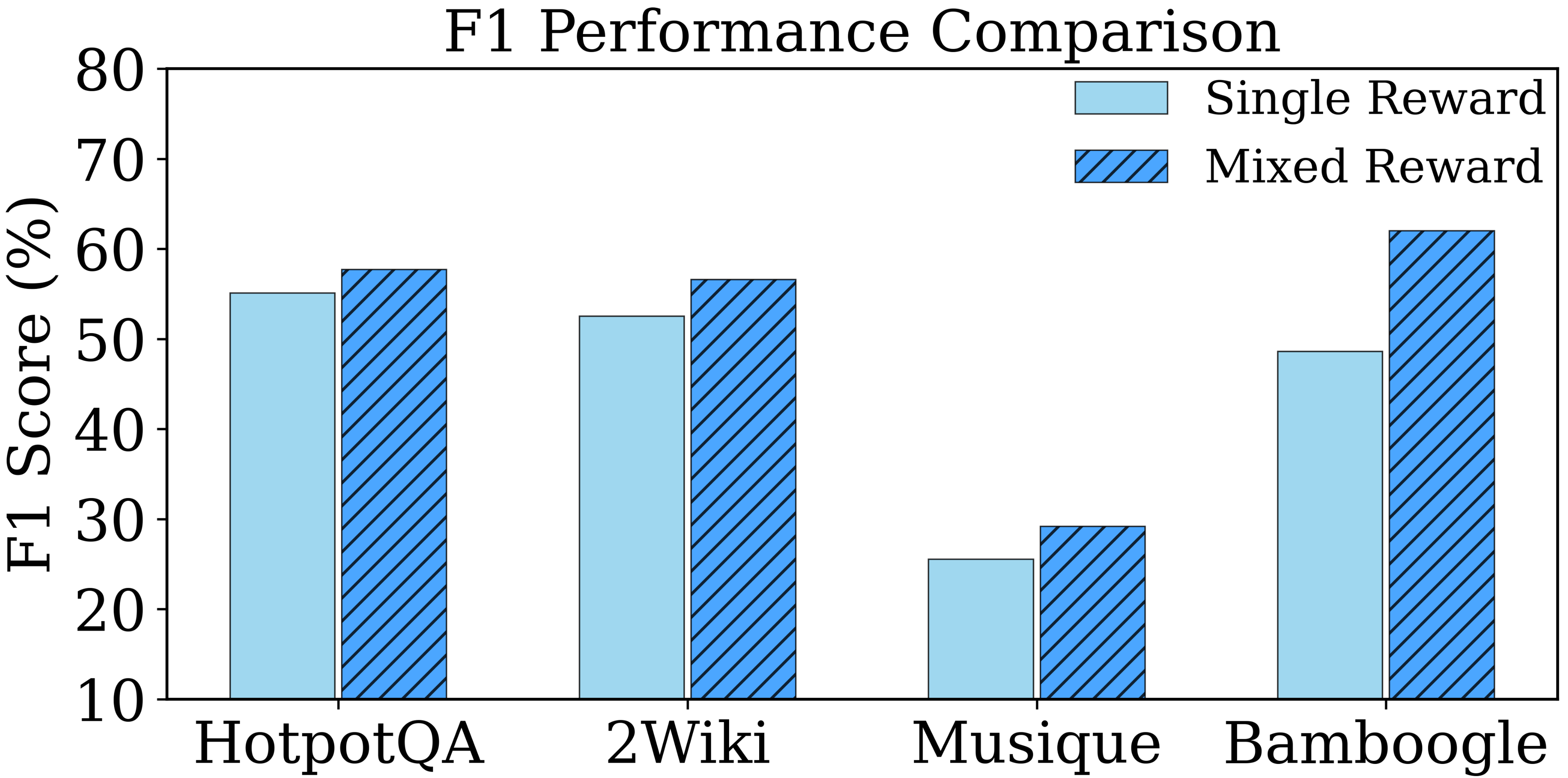}
    \caption{F1}
    \label{fig:three:b}
  \end{subfigure}%
  \begin{subfigure}[t]{0.33\textwidth}
    \centering
    \includegraphics[width=\linewidth]{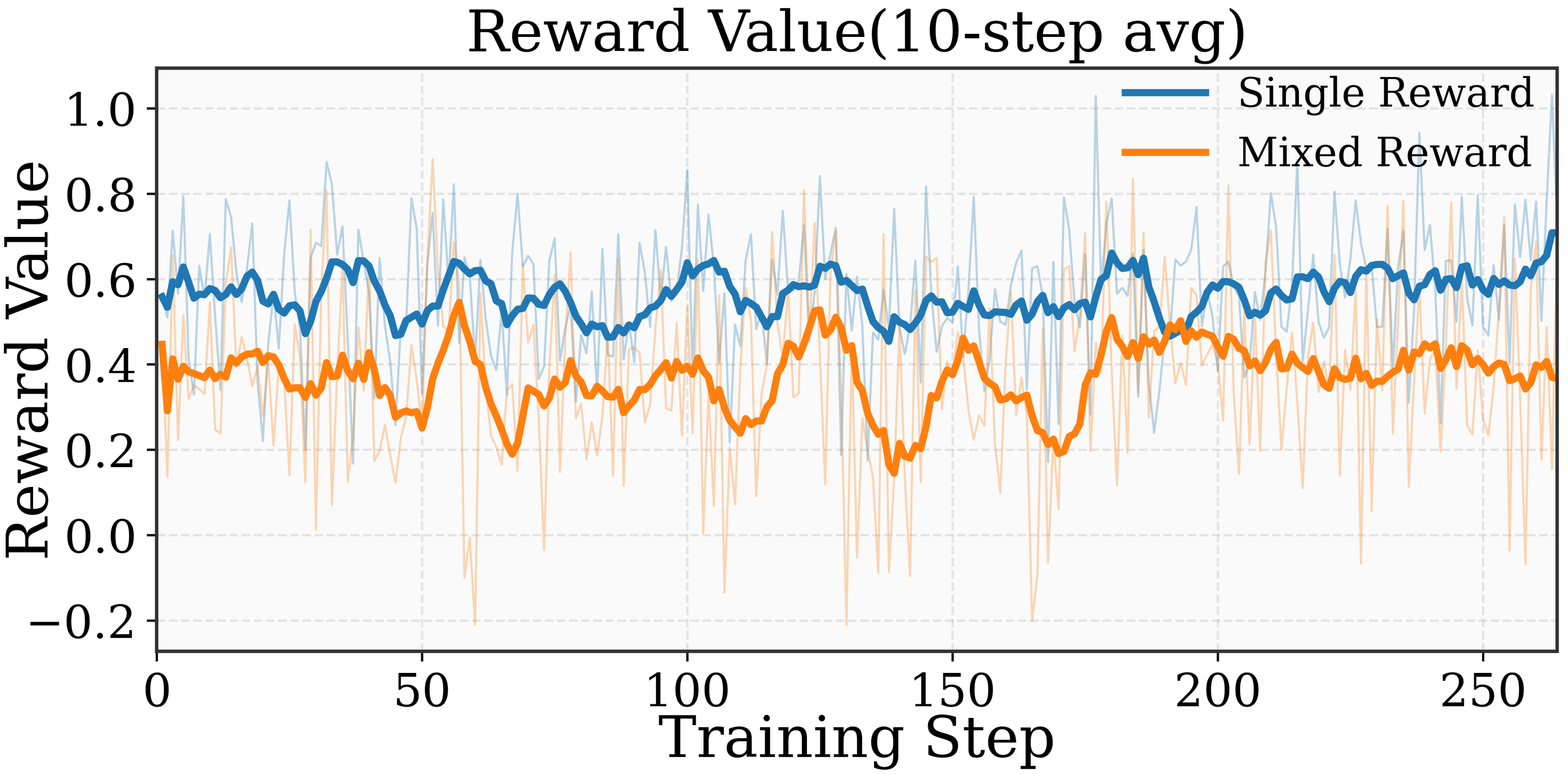}
    \caption{Reward}
    \label{fig:two}
  \end{subfigure}
  
  \caption{(a) and (b): Visualization of EM and F1 metrics for TRIO with single or mixed reward training. (c): The reward values trajectory is shown. Mixed reward design surpasses single reward design in overall performance.}
  \label{fig:reward_design}
  \vspace{-14pt}
\end{figure*}
\subsection{Ablation Study}
In order to evaluate the impact of each specific agent optimization in TRIO, we conduct ablation studies on the multi-hop benchmarks. Specifically, we evaluate three variants: (1) TRIO w/o Query Opt.: replace the trained query extension agent with the base backbone. (2) TRIO w/o Judge Opt.: replace the trained judge with the base backbone. (3) TRIO w/o Answer Opt.: replace the trained answer agent with the base backbone. The experimental results are reported in Table~\ref{tab:ablation_study}. On average, all variants perform well below TRIO, with a larger drop on out-of-domain datasets than on in-domain datasets. This indicates that each specialized agent is effective and that collaboration among the role-specific agents further improves performance, especially on unseen data. 
Specifically, TRIO w/o Answer Opt. performs worst on all benchmarks, indicating that a well-trained answer agent is critical for multi-agent RAG systems. TRIO w/o Judge Opt. is the strongest among the variants and even shows a slight EM gain over TRIO on 2Wiki. But TRIO w/o Judge Opt. still presents worse performance on out-of-domain datasets than in-domain datasets, suggesting that the supervised training method for the judge agent potentially leads to overfitting. Moreover, although all variants underperform TRIO, they still exceed the basic RAG method and even some recent RAG methods, reinforcing that specialized multi-agent designs strengthen RAG. These results suggest that targeted training enables effective coordination, thereby improving factual accuracy.

\begin{table}[t] 
\vspace{6pt}
\centering
\captionsetup{skip=0pt}

\newcolumntype{C}{>{\centering\arraybackslash}X} 

\setlength{\tabcolsep}{4pt} 
\renewcommand{\arraystretch}{1.25}
\setlength{\aboverulesep}{0pt}
\setlength{\belowrulesep}{0pt}

\begin{tabularx}{\linewidth}{l CCCC} 
    \toprule
     & \multicolumn{2}{c}{\textbf{Musique}} & \multicolumn{2}{c}{\textbf{Bamboogle}} \\
    \cmidrule(lr){2-3}\cmidrule(lr){4-5}
    \textbf{Method} & {\textbf{EM}} & {\textbf{F1}} & {\textbf{EM}} & {\textbf{F1}} \\
    \midrule
    \rowcolor{white}
    Fixed Round = 1      & 0.1556 & 0.2597 & 0.4480 & 0.5614 \\
    Fixed Round = 2      & 0.1585 & 0.2609 & 0.4480 & 0.5587 \\
    \rowcolor{white}
    Fixed Round = 3      & 0.1585 & 0.2599 & 0.4560 & 0.5811 \\
    Fixed Round = 4      & 0.1547 & 0.2561 & 0.4560 & 0.5816 \\
    \rowcolor{white}
    Fixed Round = 5      & 0.1531 & 0.2516 & 0.4560 & 0.5818 \\
    \cmidrule(l{0pt}r{0pt}){1-5}
    \rowcolor{gray!15}
    Flexible Round & 0.1857 & 0.2818 & 0.4960 & 0.6200 \\
    \bottomrule
\end{tabularx}
\vspace{3pt}
\caption{TRIO performance with flexible and different fixed iteration rounds on the retrieval step.}
\label{tab:rounds}
\vspace{-1pt}
\end{table}

\subsection{Analysis of Retrieval Rounds}

To systematically assess the impact of different retrieval rounds on the performance of TRIO. We fix the number of rounds in the retrieval step from one to five and evaluate them on the two out-of-domain multi-hop datasets. The results are shown in Table~\ref{tab:rounds}, where the flexible round is TRIO with a maximum round equal to three, and the end decision is made by the judge agent. Fixing the round count reduces performance across all variants. On the Musique dataset, as the number of retrieval rounds increases, the F1 score initially increases slightly and then declines. On the Bamboogle dataset, the F1 score shows a large jump when the round is equal to three and only minor gains thereafter. These trends suggest that longer contexts can help, but also introduce irrelevant noise, increasing the risk of hallucination. In addition, the flexible setting achieves the best results, indicating that supplying an appropriate number of retrieval chunks is beneficial and that the judge agent effectively decides when to stop and answer.

\subsection{Reward Design Analysis}
The existing reasoning RAG methods~\citep{R1Searcher,searchR1COLM} often rely on a single thought throughout the training process. But the single reward struggles with weak generalization across RAG tasks. To investigate the impact of two-stage reward design on TRIO, we train the answer agent only with a single reward $R_{a2}$ (see Equation~\ref{eq:reward2}) and the proposed two-stage reward design (described in Section~\ref{sec:answer_agent}, both using the same Llama3.1-8B-Instruct as backbone model). The results and reward values in the last 264 steps are shown in Figure~\ref{fig:reward_design}. We find that single reward design yields a small drop (-5.9\% on average) in two in-domain datasets and a much larger drop in two out-of-domain datasets (-17.03\% on average), indicating that the two-stage reward improves overall performance, especially for generalization. 
Furthermore, comparing reward values over the last 264 training steps, where the reward functions are the same for both designs, further supports this finding. The single reward method achieves higher training rewards than the mixed reward design. This pattern also suggests that the training model with a single reward is prone to overfitting the training distribution. We conclude that using the two-stage reward design to train the answer agent reduces overfitting and improves generalization to unseen data.

\subsection{Efficiency Analysis}

To investigate the real-world inference efficiency of our method, we deployed TRIO and all baselines on two L40S GPUs with identical VRAM allocation. We sampled 100 questions from each multi-hop benchmark and measured the inference time across all methods. Table~\ref{tab:inference_efficiency} presents the overall inference time of the eight retrieval-augmented baselines. While early loop-based methods, Self-RAG and Iter-RetGen, require less inference time, their performance is significantly weaker than recent reasoning and multi-agent methods, illustrating the typical trade-off between accuracy and speed. 

In contrast, TRIO-4B achieves 98\% of the performance of reasoning-based methods while requiring only half the inference time. Furthermore, TRIO-8B outperforms these reasoning baselines while using slightly less inference time. As further analyzed in Appendix~\ref{sec:case_study}, TRIO's efficiency stems from generating a highly compact number of output tokens during inference. This minimal token generation makes it highly suitable for large-scale real-world deployment while maintaining interpretability.

\begin{table}[t] 
\centering
\small
\vspace{8pt}
\setlength{\tabcolsep}{0pt}
\renewcommand{\arraystretch}{1.25}
\setlength{\aboverulesep}{0pt}
\setlength{\belowrulesep}{0pt}

\newcolumntype{M}{>{\hsize=1.28\hsize\raggedright\arraybackslash}X} 
\newcolumntype{H}{>{\hsize=1.06\hsize\centering\arraybackslash}X}   
\newcolumntype{W}{>{\hsize=0.77\hsize\centering\arraybackslash}X}   
\newcolumntype{Q}{>{\hsize=1.04\hsize\centering\arraybackslash}X}   
\newcolumntype{B}{>{\hsize=1.20\hsize\centering\arraybackslash}X}   
\newcolumntype{A}{>{\hsize=0.65\hsize\centering\arraybackslash}X}  


\begin{tabularx}{\linewidth}{M H W Q B A}
    \toprule
     & \multicolumn{5}{c}{\textbf{Inference Time (s / 100 queries)}} \\
    \cmidrule(lr){2-6}
    \textbf{Method} & {\textbf{Hotpotqa}} & {\textbf{2Wiki}} & {\textbf{Musique}} & {\textbf{Bamboogle}} & {\textbf{Avg}} \\
    \midrule
    \rowcolor{white}
    SuRe        & 603.2 & 711.8 & 521.7 & 687.4 & 631.0 \\
    IRCOT       & 64.8  & 90.1  & 84.5  & 82.7  & 80.5 \\
    \rowcolor{white}
    Iter-RetGen & 20.4  & 23.2  & 20.5  & 20.4  & 21.1 \\
    Self-RAG    & 10.9  & 10.2  & 9.6   & 10.1  & 10.2 \\
    \rowcolor{white}
    Search-R1   & 91.5  & 86.4  & 97.4  & 85.1  & 90.1 \\
    R1-Searcher & 107.4 & 104.9 & 65.6  & 74.2  & 88.0 \\
    \rowcolor{white}
    DRAG        & 571.3 & 699.4 & 852.4 & 660.1 & 695.8 \\
    \cmidrule(l{0pt}r{0pt}){1-6}
    \rowcolor{gray!15}
    TRIO-8B     & 80.4  & 82.1  & 74.9  & 82.3  & 79.9 \\
    \rowcolor{gray!15}
    TRIO-4B     & 43.6  & 61.1  & 40.8  & 36.9  & 45.6 \\
    \bottomrule
\end{tabularx}
\captionsetup{skip=2pt}
\caption{Inference time of eight methods (seconds per 100 queries) across different datasets.}
\label{tab:inference_efficiency}

\end{table}

\section{Conclusion} \label{sec:conclusion}
\vspace{-5pt}
We present TRIO, a framework that enhances RAG by coordinating three individually optimized agents for query extension, context sufficiency judgment, and answer generation. Starting from question-answer supervision only, we design distinct data generation methods for each subtask and optimize agents using reinforcement learning or supervised fine-tuning with their specific corpora. Experiments demonstrate that TRIO-8B achieves the best average score and the best result on four of the five main benchmarks. Our analysis reveals that subtask-specific training corpora and mixed reward functions enhance each agent's specialization, while effective coordination improves answer quality. TRIO can be readily extended to various scenarios and provides a new direction for incorporating multi-agent cooperation into RAG systems.

\section*{Limitations}

TRIO demonstrates strong effectiveness in improving retrieval efficiency and answer quality in different benchmarks. However, TRIO is evaluated primarily on short-answer QA and has not yet been validated for live web search or long-form generation. Its adaptive loop may fail when on tasks requiring very long-form responses or extended retrieval chains, as errors and irrelevant evidence may accumulate across retrieval rounds and hinder effective answer generation~\citep{careacl}. Live web search may introduce additional challenges, including changing document availability and heterogeneous webpage structures. Future work should calibrate retrieval-stopping decisions and evaluate these failure modes in dynamic, real-world corpora.


\section*{Acknowledgement}
This research was supported in part by the Australian Research Council (ARC) under discovery grant DP250103612 and DP260101395, ARC Research Hub for Human-Robot Teaming for Sustainable and Resilient Construction (ITRH) grant IH24010016, and Australian National Health and Medical Research Council (NHMRC) Ideas Grant APP2021183.

\bibliography{acl_reference} 

\appendix
\newcolumntype{Y}{>{\RaggedRight\arraybackslash}X}
\arrayrulecolor{black!55}          
\setlength{\arrayrulewidth}{0.4pt} 
\setlength{\tabcolsep}{6pt}
\renewcommand{\arraystretch}{1.18}

\clearpage
\newpage

\section{AI Assistants Usage}
We use large language models (ChatGPT) to refine grammar and enhance sentence clarity in our manuscript, and to aid in debugging code by explaining errors. This disclosure follows the ACL 2026 policy on AI usage.

\section{Prompts and Details of TRIO} \label{sec:prompt_TRIO}

\subsection{Inference loop for TRIO}

The multi-agent workflow begins by putting the raw question to the retriever to obtain the initial chunks from the external corpus:
\begin{equation}
Q_0, C_0 = \mathcal{R}(x,k),
\end{equation}
where $Q_0$ and $C_0$ represent the initial query pool and its corresponding set of retrieved chunks for question $x$. The retriever $\mathcal{R}$ maps a query pool $Q$ to the union of the top-k chunks returned from the corpus for each query.

In the iterative retrieval steps, each round begins with the \emph{query extension agent} $\mathcal{A}_{q}$ using the previous query pool and retrieval chunks $(Q_n, C_n)$ to identify the knowledge gaps in answering the question $x$ and generate a new query. And then send the query to the retriever, to a new query pool $Q_{n+1}$ and corresponding retrieved chunks:
\begin{equation}
Q_{n+1}, C_{n+1} = Q_{n} + \mathcal{R}(\mathcal{A}_{q}(x, Q_n, C_n)).
\end{equation}
At the end of each retrieval round, the \emph{judge agent} $\mathcal{A}_{j}$ judges whether the current query pool is sufficient to answer the question $x$:
\begin{equation}
D = \mathcal{A}_{j}(x, Q_{n+1}, C_{n+1}).
\end{equation}
If the decision $D$ is that the current information is insufficient to answer the question, we iterate to the next round(up to a fixed round budget). Otherwise, we switch to the answer step in which the \emph{answer agent} generates the final answer based on the optimized query pool and their corresponding retrieved chunks:
\begin{equation}
y = \mathcal{A}_{g}(x, Q^*, C^*),
\end{equation}
where $\mathcal{A}_{g}$ denotes the answer agent, and y is the final answer to the question $x$.

\subsection{Prompts of Inference}
The prompt below is for the query extension agent to generate exactly one new retrieval query in a multi-step retrieval process. Given the question, the existing query pool, and retrieved documents, the agent identifies missing information and proposes a keyword-based query. The output is strictly limited to at most eight keywords, with no sentences or stopwords, and must differ from all existing queries to encourage diversity and reduce redundant retrieval.

\begin{table}[h]

  \centering
  \begin{tabularx}{\linewidth}{|Y|}
  \hline
  \rowcolor{black!12}
  \textbf{Prompt for Query Extension Agent}\\
  \hline

\textbf{System Prompt:}\\
You are a retrieval query generation agent.\\
Input:\\
QUESTION: ...\\
EXISTING QUERIES and RETRIEVED DOCUMENTS: ...\\
\\
Task:\\
Generate exactly one NEW QUERY that:\\
1) Directly helps answer the QUESTION or fill gaps not covered by EXISTING QUERIES and RETRIEVED DOCUMENTS;\\
2) Is different from all EXISTING QUERIES.\\
\\
Output rules:\\
 Only keywords; no sentences, stopwords, or connectors (e.g., and, or, of, the, about).\\
 Separate keywords with a single space;\\
 Maximum 8 keywords.\\

Output:
Only the NEW QUERY text. No explanations or extra text.\\
\\
  \textbf{User Prompt:}\\
Question: \textit{|Input Question|}\\
Query Pool: \textit{|Current Query Pool|}\\
\hline

  \end{tabularx}
  \label{tab:query_extension_prompt}

\end{table}

The prompt below instructs the answer agent to produce a final answer using the retrieval context associated with the current query pool. The agent is required to output only the final answer, without explanations or additional text, ensuring a clean, comparable output format. This design supports consistent evaluation of answer correctness across different inference settings.
\newpage

\begin{table}[h]

  \centering
  \begin{tabularx}{\linewidth}{|Y|}
  \hline
  \rowcolor{black!12}
  \textbf{Prompt for Answer Agent}\\
  \hline

  \textbf{System Prompt:}\\
Answer the question based on the given document. Output only the final answer with no explanations or additional text.\\
Query Pool: \\
\textit{|Current Query Pool|}\\

  \textbf{User Prompt:}\\
Question: \textit{|Input Question|}\\
\hline

  \end{tabularx}
  \label{tab:answer_prompt}
\end{table}

The prompt below guides the judge agent in deciding whether the existing queries and retrieved documents provide sufficient information to answer the question correctly. The agent outputs only a binary decision: “Sufficient Information” or “Insufficient Information”, with no further explanation. This minimal output enables the system to control whether additional query expansion and retrieval rounds are needed.

\begin{table}[ht]
  \centering
  \begin{tabularx}{\linewidth}{|Y|}
  \hline
  \rowcolor{black!12}
  \textbf{Prompt for Judge Agent}\\
  \hline

  \textbf{System Prompt:}\\
Determine whether the EXISTING QUERIES and RETRIEVED DOCUMENTS provide sufficient information to answer the QUESTION correctly. Output only one of the following: ‘Sufficient Information’ or ‘Insufficient Information’. Do not include explanations or additional text.\\

  \textbf{User Prompt:}\\
Question: \textit{| Input Question |}\\
Query Pool: \textit{| Current Query Pool |}\\
\hline

  \end{tabularx}
  \label{tab:judge_prompt}
\end{table}

\subsection{Setting Details for TRIO Training}
All the 8148 training questions come from two multi-hop QA benchmarks: HotpotQA~\citep{hotpotqa} and 2WikiMultiHopQA~\citep{2wiki}. The answer agent and query extension agent are trained using GRPO reinforcement learning. For the judge agent, we constructed 49785 candidate samples for SFT. Among them, 5312 samples produced mixed answers and were discarded, representing approximately 10.7\%. The remaining 44473 samples were used for training, including 9087 sufficient samples and 35386 insufficient samples. Notably, we do not need any additional human annotations, and all the supervised signals were generated by the TRIO training loop (see Section~\ref{sec:method}). Each sample is rolled out 16 times under GRPO. We included more details and the trained models in the code we provided. Train batch size is 512, rollout batch size is 32, with the learning rate set to $10^{-6}$ and the KL divergence set to $10^{-4}$.  And the judge agent is trained using supervised fine-tuning. Training tasks are conducted on a single node with two H100 GPUs. For the query-extension agent, reward computation is offloaded to two additional L40 GPUs. All the evaluation experiments are conducted on two L40S GPUs.



\subsection{Prompts in Training}

The prompt below is used to train the Judge Agent to output a binary sufficiency label indicating whether the current queries and retrieved documents contain sufficient evidence to answer the question correctly. It enforces a strict output format (“Sufficient Information”/“Insufficient Information”) to enable consistent supervision.

\begin{table}[h]
  \centering
  \begin{tabularx}{\linewidth}{|Y|}
  \hline
  \rowcolor{black!12}
  \textbf{Prompt for Judge Agent in Training}\\
  \hline

\textbf{System Prompt:}\\
Determine whether the EXISTING QUERIES and RETRIEVED DOCUMENTS provide sufficient information to answer the QUESTION correctly. Output only one of the following: ‘Sufficient Information’ or ‘Insufficient Information’. Do not include explanations or additional text.\\
\textbf{User Prompt:}\\
Question: Are the directors of both films I Can Do Bad All By Myself (Film) and Shit Year from the same country?\\
EXISTING QUERIES and RETRIEVED DOCUMENTS\\
Query 1: Are the directors of both films I Can Do Bad All By Myself (Film) and Shit Year from the same country?\\
Retrieved Content:\\
Doc 1: ...\\
Doc 2: ...\\
Doc 3: ...\\
Doc 4: ...\\
Doc 5: ...\\

\hline

  \end{tabularx}
  \label{tab:judge_prompt_t}
\end{table}

The prompt below is used to train the Query Extension Agent to produce one additional keyword query based on the question and the current retrieval results. The output is constrained to a short, non-overlapping query to encourage informative and diverse retrieval during training.

\newpage

\begin{table}[t]

  \centering
  \begin{tabularx}{\linewidth}{|Y|}
  \hline
  \rowcolor{black!12}
  \textbf{Prompt for Query Extension Agent in Training }\\
  \hline

\textbf{System Prompt:} \\
You are a retrieval query generation agent.\\
Input:\\
QUESTION: ...\\
EXISTING QUERIES and RETRIEVED DOCUMENTS: ...\\
\\
Task:\\
Generate exactly one NEW QUERY that:\\
1) Directly helps answer the QUESTION or fill gaps not covered by EXISTING QUERIES and RETRIEVED DOCUMENTS;\\
2) Is different from all EXISTING QUERIES.\\
\\
Output rules:\\
 Only keywords; no sentences, stopwords, or connectors (e.g., and, or, of, the, about).\\
 Separate keywords with a single space;\\
 Maximum 8 keywords.\\

Output:
Only the NEW QUERY text. No explanations or extra text.\\
\\
\textbf{User Prompt:}\\
Question: Are the directors of both films I Can Do Bad All By Myself (Film) and Shit Year from the same country?\\
EXISTING QUERIES and RETRIEVED DOCUMENTS\\
Query 1: Are the directors of both films I Can Do Bad All By Myself (Film) and Shit Year from the same country?\\
Retrieved Content:\\
Doc 1: ...\\
Doc 2: ...\\
Doc 3: ...\\
Doc 4: ...\\
Doc 5: ...\\

\hline

\end{tabularx}
  \label{tab:query_extension_prompt_t}
\end{table}


The prompt below is used to train the Answer Agent to produce the final answer from the provided multi-query retrieval context. It presents the question along with the existing queries and their retrieved documents, and enforces an answer-only output without explanations to maintain consistency in supervision and evaluation.

\begin{table}[t]
  \centering
  \begin{tabularx}{\linewidth}{|Y|}
  \hline
  \rowcolor{black!12}
  \textbf{Prompt for Answer Agent in Training}\\
  \hline

\textbf{System Prompt:}\\
Answer the question based on the given document. Output only the final answer with no explanations or additional text.\\
EXISTING QUERIES and RETRIEVED DOCUMENTS\\
Query 1: Are the directors of both films I Can Do Bad All By Myself (Film) and Shit Year from the same country?\\
Retrieved Content:\\
Doc 1(Title: "I Can Do Bad All by Myself (film) "):  ...\\
Doc 2(Title: "I Can Do Bad All by Myself (film) ") : ...\\
Doc 3(Title: "Shit Year") : ...\\
Doc 4(Title: "Shit Year"): ...\\
Doc 5(Title: Themselves): ...\\

Query 2: Tyler Perry\\
Retrieved Content:\\
Doc 1(Title: "Tyler Perry"): ...\\
Doc 2(Title: "Tyler Perry Studios"): ... \\
Doc 3(Title: "Tyler Perry"): ... \\
Doc 4(Title: "Tyler Perry Studios"): ... \\
Doc 5(Title: "Tyler Perry"): ... \\

\textbf{User Prompt:}\\
Question: Are the directors of both films I Can Do Bad All By Myself (Film) and Shit Year from the same country?\\

\hline

  \end{tabularx}
  \label{tab:answer_prompt_p}
\end{table}

\subsection{Comparison with Existing Adaptive RAG Methods}

Existing adaptive RAG methods differ in where adaptation is introduced. RQ-RAG~\citep{rqrag} trains a query-refinement model. IRCoT~\citep{IRCOT} interleaves prompting-based reasoning and retrieval. Self-RAG~\citep{asai2024selfrag} trains a single generator to emit reflection tokens. Search-R1~\citep{searchR1COLM} and R1-Searcher~\citep{R1Searcher} optimize a single retrieval-reasoning policy. DRAG~\citep{drag} and MA-RAG~\citep{marag} coordinate role-specialized agents mainly through debate or collaborative prompting without task-specific training. In contrast, TRIO sequentially optimizes separate query-extension, stopping, and answer agents, with answer agent outcomes providing supervision for query extension and sufficiency judgment.

\clearpage

\section{Prompts of Supervised Data Generation} \label{sec:data_generation}

The prompts below implement a lightweight retrieval debate mechanism to assess the sufficiency of evidence. The Proponent argues that retrieved documents support an answer and provide a conclusion without requesting further retrieval, while the Opponent challenges sufficiency and either proposes a new query or concedes if evidence is complete. The Moderator then selects the more convincing argument and outputs only the winning agent name to drive the pipeline.

\begin{table}[ht]
  \centering
  \begin{tabularx}{\linewidth}{|Y|}
  \hline
  \rowcolor{black!12}
  \textbf{Prompt for Moderator in Retrieval Debate}\\
  \hline

  \textbf{System Prompt:}\\
You are the judge in a debate. Your task is to evaluate the arguments from agents. There are two types of agents:\\
1. Proponent Agent: Argues that the current retrieved content is sufficient.\\
2. Opponent Agent: Argues that the current retrieved content is insufficient and proposes query refinement.\\
\\
  \textbf{User Prompt:}\\
Question:  \textit{|Input Question|}\\
Query Pool:  \textit{|Current Query Pool|}\\
Proponent:\\
 \textit{|Output of Proponent|}\\
Opponent:\\
 \textit{|Output of Opponent|}\\
\\
Output only the agent's name. Do NOT output more than one agent or any explanation.\\
\hline

\end{tabularx}
  \label{tab:moderator_prompt}
\end{table}

\begin{table}[ht]
  \centering
  \begin{tabularx}{\linewidth}{|Y|}
  \hline
  \rowcolor{black!12}
  \textbf{Prompt for Proponent in Retrieval Debate}\\
  \hline

  \textbf{System Prompt:}\\
You are a debater. Argue that the current retrieved content is sufficient to answer the question and try to give the answer based on the given documents. Deliver a brief, strong argument with clear reasoning. Do not suggest further retrieval. No extra explanations.\\
\\
  \textbf{User Prompt:}\\
  \hspace*{2em}Question: \textit{|Input Question|}\\
  \hspace*{2em}Query Pool: \textit{|Current Query Pool|}\\
\hline

  \end{tabularx}
  \label{tab:Proponent_prompt}
\end{table}

\begin{table}[ht]
  \centering
  \begin{tabularx}{\linewidth}{|Y|}
  \hline
  \rowcolor{black!12}
  \textbf{Prompt for Opponent in Retrieval Debate}\\
  \hline

\textbf{System Prompt:}\\
You are a critical thinker and debater. Your task is to challenge the sufficiency of the current documents. However, your ultimate goal is to find the correct answer efficiently. If you believe the provided information is TRULY and COMPLETELY sufficient, your duty is to concede.\\
\\
 The action you can choose:\\
 1. Query Expansion: If a completely new line of inquiry is needed. 
     Output exactly in this format in the end of your response: 
     Query Expansion: [New Query]\\
 2. Concede: If the current information is truly sufficient. 
     Only Output: Concede\\
 Deliver a brief, strong argument, then you must choose one action in the exact format required.\\
 \\
  \textbf{User Prompt:}\\
Question: \textit{|Input Question|}\\
Query Pool: \textit{|Current Query Pool|}\\
\hline

  \end{tabularx}
  \label{tab:opponent_prompt}

\end{table}


\section{Case Study} \label{sec:case_study}

We provide two typical cases to show how our agents collaborate to effectively answer multi-hop questions.
Table \ref{tab:case_study_compact} illustrates how TRIO dynamically adapts to context quality. Case 1 shows an efficient "early stop" in which the Judge Agent correctly recognizes a perfect initial retrieval, avoiding additional retrieval costs. In contrast, Case 2 highlights the "Correction Pattern" for complex queries. When the initial retrieval returns related entities (e.g., "FC Barcelona") but lacks the specific league rules, the Judge Agent issues an "Insufficient" verdict. This triggers the Query Extension Agent to pivot the search towards "La Liga". This multi-round collaboration ensures that the Answer Agent generates a response once the semantic gap is bridged, effectively reducing hallucinations caused by irrelevant context. Furthermore, the entire inference process generates only 5 to 7 words. Even for complex questions that require up to three retrieval rounds, the output length remains significantly shorter than that of state-of-the-art reasoning methods, which typically generate hundreds of words per question.
\definecolor{triogray}{RGB}{240,240,240} 
\definecolor{triogreen}{RGB}{34, 139, 34}
\definecolor{trioagent}{RGB}{0, 0, 128}
\definecolor{linegray}{gray}{0.85} 

\begin{table}[!t]
\vspace{3pt}
    \centering
    \renewcommand{\arraystretch}{1.1} 
    \setlength{\tabcolsep}{3pt} 

    \begin{tabularx}{\linewidth}{X}
        \toprule[1.5pt]
        
        \textbf{Case 1: Immediate Success (Early Stop)} \\ 
        \arrayrulecolor{linegray}\hline\arrayrulecolor{black}
        \textbf{Question:} Who played the character in the Santa Clause 3 that has a series named after it that includes Frost at Christmas? \\
        \textbf{Ground Truth:} Martin Short \\
        
        \arrayrulecolor{linegray}\hline\arrayrulecolor{black} 
        
        \textbf{Round 0:} \\
        \textbf{Retrieval Query:} Who played the character in the Santa Clause 3 that has a series named after it that includes Frost at Christmas? \\
        \textbf{Retrieved Content:}
        1. \textit{The Santa Clause 3}; 2. \textit{The Santa Clause (film series)}; 3. \textit{The Santa Clause 3: The Escape Clause}... \\

        \textbf{\textcolor{trioagent}{Judge:}} Sufficient \textcolor{gray}{\textit{(Decision: Stop \& Answer)}} \\
        \arrayrulecolor{linegray}\hline\arrayrulecolor{black}
        \textbf{\textcolor{trioagent}{Answer:}} Jack Frost played by Martin Short. \\
        
        \midrule[1pt] 
        
        \textbf{Case 2: Iterative Refinement (Correction)} \\
        \arrayrulecolor{linegray}\hline\arrayrulecolor{black}
        \textbf{Question:} How many games do team in the league where Barcelona won titles in 1948 and 1949 play? \\
        \textbf{Ground Truth:} 38 \\
        
        \arrayrulecolor{linegray}\hline\arrayrulecolor{black}
        
        \textbf{Round 0: }\\
        \textbf{Retrieval Query:} How many games do team in the league where Barcelona won titles in 1948 and 1949 play? \\
        \textbf{Retrieved Content:}
        1. \textit{UER Pineda de Mar}; 2. \textit{1949 Campeonato Profesional}; 3. \textit{History of FC Barcelona}... \\
        \textbf{\textcolor{trioagent}{Judge:}} Insufficient \textcolor{gray}{\textit{(Reason: Lacking league rules.)}} \\
        \arrayrulecolor{linegray}\hline\arrayrulecolor{black}

        \textbf{Round 1:} \\
        \textcolor{trioagent}{\textbf{Query Extension}}: ``La Liga'' \\
        \textbf{Retrieval Query:} La Liga \\
        \textbf{Retrieved Content:} 1. \textit{La Liga}; 2. \textit{Liga de Futbol Profesional}... \\

         \textbf{\textcolor{trioagent}{Judge:}} Sufficient \textcolor{gray}{\textit{(Decision: Gap filled.)}} \\
         \arrayrulecolor{linegray}\hline\arrayrulecolor{black}
        \textbf{\textcolor{trioagent}{Answer:}} 38 \\
        
        \bottomrule[1.5pt]
    \end{tabularx}
    \caption{Case studies demonstrating the token efficiency of TRIO. Driven by the coordination among the three agents, the system completes tasks with highly concise token outputs. }
    \label{tab:case_study_compact}

\end{table}

\section{Resource Analysis}

To further investigate the training resource requirements of TRIO, we report the computational cost and memory usage of TRIO-8B and TRIO-4B in Table~\ref{tab:training_resources}. As a reference point, a previous work~\citep{train_time} reports that its reproduction with Qwen2.5-3B requires 14.7 hours on 4 H200 GPUs, corresponding to 58.8 H200 GPU-hours. In comparison, our TRIO-4B version requires 58.1 H100 GPU-hours, which is slightly lower in absolute GPU-hour cost despite training multiple specialized agents and using larger models. In addition, the peak training memory usage is approximately 94 GB per GPU, or 188 GB in aggregate across two GPUs. At deployment time, current TRIO uses three independently optimized model checkpoints. The total deployment memory footprint is approximately 40 GB for TRIO-4B and 60 GB for TRIO-8B. For training, the answer agent is optimized with GRPO on 18380 samples constructed from 8148 questions, the query extension agent is trained with GRPO on 10000 samples, and the judge agent is trained with supervised fine-tuning on 44473 samples.

\begin{table}[t]
\centering
\setlength{\tabcolsep}{4pt}
\renewcommand{\arraystretch}{1.08}
\begin{tabular}{@{}lcc@{}}
\toprule
\textbf{Agent}
& \textbf{Stage}
& \shortstack{\textbf{GPU-hours}\\\textbf{(H100)}} \\
\midrule

\multicolumn{3}{@{}l}{TRIO-8B (Llama-8B)} \\
Answer agent          & Stage 1 & 33.38h \\
Answer agent          & Stage 2 & 13.06h \\
Query extension agent & --      & 33.96h \\
Judge agent           & --      & 12.60h \\
\multicolumn{2}{@{}l}{Total}    & 93.00h \\

\midrule
\multicolumn{3}{@{}l}{TRIO-4B (Qwen-4B)} \\
Answer agent          & Stage 1 & 21.80h \\
Answer agent          & Stage 2 & 8.87h \\
Query extension agent & --      & 20.60h \\
Judge agent           & --      & 6.83h \\
\multicolumn{2}{@{}l}{Total}    & 58.10h \\

\bottomrule
\end{tabular}
\caption{Training resource consumption of TRIO-8B and TRIO-4B.}
\label{tab:training_resources}
\end{table}

\section{Additional Results and Robustness Analysis}
\label{sec:noise_robustness}

To further investigate TRIO’s robustness against distracting retrieval contexts, we conduct an additional noisy retrieval experiment. Specifically, we introduce two noise levels by replacing chunks of the retrieved context with rank 998-1000 noisy chunks from the corpus. In the low-noise setting, one chunk among the top-5 retrieved chunks is replaced, while in the high-noise setting, three are replaced. The results are shown in Table~\ref{tab:noisy_retrieval}. 

TRIO demonstrates greater robustness to retrieval noise than existing methods. For example, under the high-noise setting, TRIO-8B maintains EM scores of 0.1348 and 0.4000 on MuSiQue and Bamboogle, respectively, while Search-R1 decreases substantially to 0.0144 and 0.1600. 
Interestingly, we observe that reasoning-based Search-R1 is more sensitive to noisy retrieval contexts, indicating that stronger reasoning ability alone does not necessarily guarantee robustness when the retrieved context contains highly irrelevant information. In contrast, TRIO benefits from the collaboration among the query extension, sufficiency judgment, and answer agents, enabling the system to better identify and utilize useful evidence while reducing the influence of distracting contexts. We also report the F1 scores of compared methods and our methods across the four benchmarks in Table~\ref{tab:additional_f1}.

\begin{table}[t]
\centering
\small
\setlength{\tabcolsep}{6pt}
\renewcommand{\arraystretch}{1.1}
\begin{tabular}{@{}lcccc@{}}
\toprule
& \multicolumn{2}{c}{\textbf{MuSiQue}}
& \multicolumn{2}{c}{\textbf{Bamboogle}} \\
\cmidrule(lr){2-3}
\cmidrule(lr){4-5}
\textbf{Method}
& \textbf{Low}
& \textbf{High}
& \textbf{Low}
& \textbf{High} \\
\midrule
SuRe        & 0.0645 & 0.0508 & 0.1280 & 0.1200 \\
IRCoT       & 0.0662 & 0.0559 & 0.1520 & 0.1920 \\
Iter-RetGen & 0.0699 & 0.0488 & 0.1760 & 0.1600 \\
DRAG        & 0.1129 & 0.1067 & 0.2720 & 0.2640 \\
Self-RAG    & 0.0384 & 0.0372 & 0.0640 & 0.0560 \\
Search-R1   & 0.1531 & 0.0144 & 0.1920 & 0.1600 \\
TRIO-4B     & 0.1407 & 0.1283 & 0.3440 & 0.3120 \\
\textbf{TRIO-8B}
            & \textbf{0.1708}
            & \textbf{0.1348}
            & \textbf{0.4560}
            & \textbf{0.4000} \\
\bottomrule
\end{tabular}
\caption{EM scores under low and high noise settings. Low and High denote replacing one and three of the top-5 retrieved chunks with noisy chunks, respectively.}
\label{tab:noisy_retrieval}
\end{table}

\begin{table}[t]
\centering
\small
\setlength{\tabcolsep}{2.8pt}
\renewcommand{\arraystretch}{1.1}
\begin{tabular}{@{}lcccc@{}}
\toprule
\textbf{Method}
& \textbf{HotpotQA}
& \textbf{2Wiki}
& \textbf{MuSiQue}
& \textbf{Bamboogle} \\
\midrule
SuRe
& 0.3538 & 0.1917 & 0.1296 & 0.2727 \\

IRCoT
& 0.2825 & 0.2363 & 0.1244 & 0.3112 \\

Iter-RetGen
& 0.4419 & 0.2636 & 0.1297 & 0.2701 \\

DRAG
& 0.4146 & 0.2791 & 0.2010 & 0.4030 \\

Self-RAG
& 0.2842 & 0.2351 & 0.1195 & 0.1576 \\

Search-R1
& 0.5306 & 0.4879 & 0.2608 & 0.5780 \\

R1-Searcher
& 0.5200 & 0.4481 & \underline{0.2617}
& \underline{0.5782} \\

TRIO-4B
& \underline{0.5455}
& \underline{0.5023}
& 0.2555
& 0.5555 \\

TRIO-8B
& \textbf{0.5770}
& \textbf{0.5660}
& \textbf{0.2818}
& \textbf{0.6200} \\
\bottomrule
\end{tabular}
\caption{F1 results on four multi-hop benchmarks.
The best result in each column is shown in bold, and
the second-best result is underlined.}
\label{tab:additional_f1}
\end{table}

\section{Evaluation on Biomedical Domain Dataset}

To further investigate TRIO's generalization to out-of-domain questions that differ substantially from the training data, we conduct an additional evaluation on the PubMedQA benchmark~\citep{pubmedqa}. PubMedQA represents a completely unseen biomedical domain relative to the training datasets. We construct the evaluation set using the corresponding 270K biomedical retrieval corpus and evaluate all methods under the same retrieval and generation settings as the main experiments. The results are shown in Table~\ref{tab:pubmedqa}.

\begin{table}[h]
\centering
\small
\setlength{\tabcolsep}{10pt}
\renewcommand{\arraystretch}{1.1}
\begin{tabular}{@{}lc@{}}
\toprule
\textbf{Method} & \textbf{PubMedQA} \\
\midrule
SuRe        & 0.256 \\
IRCoT       & 0.458 \\
Iter-RetGen & 0.716 \\
DRAG        & 0.644 \\
Self-RAG    & 0.628 \\
Search-R1   & 0.568 \\
TRIO-4B     & 0.666 \\
TRIO-8B     & \textbf{0.766} \\
\bottomrule
\end{tabular}
\caption{Results of TRIO and other baselines on PubMedQA benchmark.}
\label{tab:pubmedqa}
\end{table}

TRIO-8B achieves the best performance on the unseen biomedical benchmark, and TRIO-4B achieves 0.666, suggesting that TRIO’s task-specific optimization can generalize beyond Wikipedia-based multi-hop QA scenarios. In addition, we observe that the reasoning-based method Search-R1 does not transfer effectively to the unseen biomedical domains. In contrast, methods using untrained general models, such as Iter-RetGen and DRAG, exhibit stronger transferability on the biomedical dataset, achieving 0.716 and 0.644 in accuracy, respectively. We also observe that the earlier method SuRe exhibits relatively limited generalization on the biomedical benchmark.

\end{document}